\documentclass{article}
\begin{document}
\title{{\bf Chatbots and Zero Sales Resistance}}
\author{Sauro Succi\\
Italian Institute of Technology,\\ 
Viale Regina Elena, 291, 00161, Rome, Italy\\
Physics Department, Harvard University, Cambridge USA
}
\maketitle

\begin{abstract}
It is argued that the pursuit of an ever increasing 
number of weights in large-scale machine learning 
applications, besides being energetically 
unsustainable, is also conducive to manipulative 
strategies whereby Science is easily served as a  
strawman for economic and financial power.  
If machine learning is meant to serve science ahead of 
vested business interests, a paradigm shift is needed: 
from more weights and little insight to more 
insight and less weights.

\end{abstract}

\section{Introduction}

Not a day goes by without we hear of the latest AI breakthroughs,
such as chatbots that write up texts or generate images increasingly 
harder to tell apart from their human-made counterparts.
These headlines come with a heavy load of hype, but even 
with hype factored out, a highly seductive promise 
stands tall, the promise
to capture levels of complexity largely out of grasp 
for our best theories, models and simulations.
Briefly, AI would supplant the time-honored Scientific 
Method, as we know it since Galileo's time \cite{WIRED,PEDRO}.

While heavily pumped up, this promise is not empty, addressing 
as it does, among others, one of the most 
vexing Achille's heels of the scientific method, the 
infamous Curse of Dimensionality (CoD) \cite{COD}.
Indeed, CoD compounds with a profound hallmark 
of Complexity, namely the fact that complex systems 
are {\it sneaky}: they inhabit ultra-dimensional spaces 
but don't fill them up \cite{OUP22,NAT,ESS}.
To the contrary, "interesting things" take place in 
ultrathin and often highly scattered portions of the 
huge state space available to them.
Nature likes to play hide and seek and big time so. 
An illuminating example can be found in the book of Frenkel and Smit \cite{FS},
where we learn that the chance of making a sensible Monte Carlo move in the
state space of hundred hard-spheres (please note, hundred, not Avogadro's) 
is about $10^{-260}$! The golden nuggets are well hidden indeed.

Computational science has devised a number of clever 
techniques to visit the regions hosting the preciously 
rare golden nuggets without waiting many ages 
of the Universe \cite{META}.
Yet, the CoD still remains a very tough cookie 
for the scientific method to the present day.

Artificial Intelligence, and more specifically 
Machine Learning (ML), promise a new and unprecedently
powerful angle of attack to Complexity.
And again, the promise is largely overblown but 
not empty, as witnessed by a number
of success stories: chess and GO winnings, self-driving cars, 
DeepFold mapping of protein structure, stand out 
as some the most spectular(ized) 
cases in point \cite{DEEP}.

\section{The basic ML procedure}

It is worth discussing where this "magic" comes from 
in some little more detail.

The basic idea of ML is to represent a given $D$-dimensional 
output $y$ (target) through the recursive application 
of a simple non linear map \cite{ML}.
For a neural network (NN) consisting of an input layer $x$,  $L$ hidden layers
$z_1 \dots z_L$, each containing $N$ neurons, and an output layer $y$,
the update chain $x \to z_1 \dots \to z_L \to  y$
reads symbolically as follows:
\begin{equation}
\label{ML}
z_1 = f(W_1 x       -b_1),\; \dots  z_L = f(W_L z_{L-1} -b_L),\;
y = f(W_{L+1} z_l -b_{L+1})
\end{equation}
where $W_l$ are $N \times N$ matrices of weights, $b_l$ are N-dimensional arrays
of biases and $f$ is a nonlinear activation function, to be chosen out 
of a large palette.
The output $y$ is then compared with a given training 
target $y_T$ (Truth) and the weights are
recursively updated in such a way as to minimize 
the discrepancy between $y$ and $y_T$ (Loss function), up 
to the desired tolerance.  
In equations
\begin{equation}
\label{LOSS}
\mathcal{L}[W] = dis (y[W],y_T) \le \epsilon
\end{equation}
where the loss function $\mathcal{L}$, namely the distance between
the model output and the target in some suitable metric, is
an explicit function of the weights configuration $W$.
The central engine of the ML procedure is an update schedule
of the weights whereby the loss function is taken below 
the desired tolerance $\epsilon$. 
This is typically obtained through some form of 
steepest descent (SD) search:

\begin{equation}
\label{SD}
W(t+1) = W(t) - \gamma \frac{\partial \mathcal{L}}{\partial W}
\end{equation}

where $t$ denotes the iteration step and 
$\gamma$ is a suitable relaxation rate.

The bold idea is that with {\it big enough} data for training, 
the ML sequence (1-2-3) can reach {\it any} target, whence 
the alleged demise of the  scientific method \cite{WIRED,PEDRO}.

Where does such magic come from?

The key point is that for a DNN (deep neural net) of depth $L$
(number of layers) and width $N$ (numbers of neurons
per layer), there are $N_P=N^L$ possible paths connecting any 
single item $x_i, \; i=1,N$ in the input layer to any another 
single item $y_j, \;j=1,N$ in the output layer.
Hence a DNN with $N=10^3$ neurons and $L=10^2$ layers features
$N_W = N^2 L=10^8$ weights and $N_P=10^{30}$ paths.
Such gargantuan network of paths represents the state-space
of the DNN learning process, which can proceed through several
concurrent paths at a time. 
If you think that this is sci-fi, please think again, as current 
leading edge ML applications, such as DeepFold or 
Large Language Models motoring the most powerful "ask-me-anything" chatbots
are using up to 100 billions weights, basically the number
of neurons in our brain. Except that our brain works at 20 Watt 
while the largest ML models are now sucking up at least 
ten million times more, a point to which we shall return shortly.

These numbers unveil the magic behind ML: DNN duel the 
CoD face up, by unleashing an exponential number of paths, and
adjusting them in such a way as to sensibly populate the 
sneaky regions where the golden nuggets are to be found.

This strategy is an opinion splitter: AI pragmatists 
are enthusiastic at the
conceptual simplicity of this black-box and, 
speaking of weights, to them 
"too much is not enough". 
Scientists fond of Insight long ahead of Control, are 
horrified at the diverging number of parameters, 
their "prejudice" being that parameters are fudge factors
concealing lack of understanding, so that their motto
is rather "the least the best".

\section{Criticism to ML}

There are indeed several solid reasons to second-guess
the blue-sky scenarios portrayed by the
most enthusisatic (and aggressive ) AI aficionados.
Here we briefly summarize the main items, some of which
have been pointed out before \cite{PVC,SC}:

1) {\it Convergence}: the convergence of the ML procedure depends
on the landscape of the Loss function in weight space, meaning
that the SD procedure may get trapped in false local minima.
This is particularly true in the case of 
hyperdimensional weight spaces. 

2) {\it Accuracy}: in order to achieve the required level of accuracy
enormous data sets might be required.
This is because uncertainty in complex systems is far more resilient
than the standard gaussian rule $1/\sqrt{N}$, $N$ being the size
of the data set. The reason is physical: the gaussian rule applies to
uncorrelated systems, whereas complex systems are typically
highly correlated, in both space and time.

3) {\it Explainability}: the optimal weights, $W_{opt}$, 
associated to a given solution, do not lend themselves to any 
direct physical interpretation. 
In other words, they provide no Insight
into the phenomenon under study.

4) {\it Causality}: the ML procedure is built upon the idea
of catching Correlations between the model solution $y=f(x,W)$
and the target $y_T$. As is well known, this says nothing about the 
causal relations that underlie such correlation.
Once again, Control without Insight.
Moreover it is well known that the ratio between false and true
correlations is an exponential function of the 
size of the dataset \cite{LONGO}.

5) {\it Uncertainty Quantification}: once an optimal weight configuration
$W_{opt}$ is found, it is extremely hard to estimate the robustness
of the solution toward changes of the weight parameters.     
The main hurdle is again the hyperdimensionality of weight space.

6) {\it Smoothness}: the activation functions commonly used
in ML are differentiable. While expedient to the simplicity
of the SD procedure, this hampers convergence in case
the output is a rough (non-differentiable) function of the input,
as it is often the case in complex systems. 

7) {\it Sustainability}: Training does not come for free: it
is estimated that new generations of chatbots with trillions
of weights may reach up to Gigawatts of power demand.

The above list of items is open for a healthy 
and hopefully hype-free scientific debate.
Yet, it appears hard to argue against the idea
that a blind escalation of the number of weights 
is hardly the best way to go, on both scientific
and societal accounts.

In the following we focus on latter, with special
emphasis on issue 7) above, which is clearly
tightly intertwined with all other items in the list. 

\section{Sustainability}

Besides the exponential number of parameters, in order to 
converge, the ML procedure needs correspondingly huge
training datasets. As mentioned above, training
does not come for free: it is estimated that new generation 
chatbots will come near to 
the Gigawatt power demand, in excess of most existing power plants.
The comparison with the 20 Watts of our brain is embarassing.
True, once trained, their routine operation is much less
expensive. The question still remains, as to whether, chatbots
can still function accurately and reliably on unseen data.

Given the energy-devouring nature of the 
large-scale ML procedure, the obvious 
question is: is it really worth sacrificing a substantial 
share of the total energy budget worldwide to the totem of chatbots?    

This may make Jensen Huang, the founder of Nvidia, the richest man on
Earth, but still it is not the way to go for the 
rest of us, apart the super-elite who may be able to afford 
extra-terrestrial life for the decades to come.
The point is that, even when it works, ML is hardly Explainable,
it offers little clue on the physical meaning of 
the parameters: Control but little Insight, if any at all.
And with Insight out of the game, there's no guarantee that what works
for seen data will keep working again for the unseen ones (extrapolation).

Of course, one can close eyes and keep going on steroids with weights, but,
given the power bill discussed above, this sounds reckless at best. 
There must be better ways. An interesting clue in this direction is
that the overwhelming majority of the weights, the experts tell me, are
close to zero, meaning that they do not contribute significantly to the 
success of the ordeal, hence offering little return on computing investment.
This is scientifically {\it very} interesting 
and it begs for understanding, not just
because this is what science is all about, but also because 
Understanding here means saving oceans of Gigawatts.   

Recently Elon Musk advocated the need of putting 
AI on a rigorous scientific basis, in his own words 
"Join xAI (Explainable AI) if you believe in our mission of 
understanding the universe, which requires maximally rigorous pursuit 
of the truth, without regard to popularity or political correctness.” 
Yann LeCun, one of the most respected computer scientists worldwide 
and Turing awardee, promptly countered that
"Musk wants a maximally rigorous pursuit of the truth but 
spews crazy-ass conspiracy theories on his own social platform.” 
And, upon being questioned by Musk about his recent science, 
LeCun goes on by quoting his some 80 papers, as
opposed to Musk's zero entries in this ballgame.
 
Now, it is ironic enough to hear one of the most muscular and hungriest
AI energy consumers on the planet to advocate the 
rigorous pursuit of scientific truth.
And despite his towering status in computer science,
it is only slightly less ironic to see
LeCun taking on the role of the guardian of science, 
given that LeCun, incidentally also  
Chief AI Scientist at Zuckerberg's Facebook, is a champion
of that kind of computer science where the dismissal of Insight
in favor of Control is largely tolerated, see 
\cite{PVCNature,PVC2024}. 
 
But let's give Musk the benefit of doubt and assume 
he's genuinely interested in understanding how the Universe works.
The tip is fairly easy: stop leveraging the 
muscular power of ML with legions of GPU's and {\it surprise us
with more understanding and less weights}.
The name of the game being causal AI or explainable AI,
acknowledging that, despite the wildest claims of AI
aficionados, whenever Correlation can replace Causality, what we
are talking about is not Science but Control.
 
This is the biggest lie which has been served to us by AAI,
where AAI stands for Aggressive AI, in order to distinguish it from the
many important contributions of AI to technology and society.
But the king is now naked, for the pursuit of Control regardless
of Insight comes with a energy price tag that 
planet Earth just cannot sustain.
Incidentally, scientists (most physicists and mathematicians) much less
visible than Musk and LeCun, are already doing this, out of the
limelight \cite{CAI1,CAI2,CAI3}.

\section{Zero sales resistance}

Musk's glorious statement about xAI calls
for another comment, mostly related to a strategy
that, back in 1944 in his prophetic book 
"The abolition of man", CS Lewis called
"Zero Sales Resistance", ZSR for short \cite{LEW}.
Pursuing financial/economic interests under the glorified
veil of world-saving intentions is a well-known strategy since 
long: what is new, though, is the unprecedented power of modern AAI.
Here comes the point.
The speed of modern technology has basically collapsed spacetime: a single
message can reach any (connected) individual on this planet in virtually no time.
The result is that a well-crafted message, suitably conveyed by your
most seductive influencer of choice (maybe a chatbot?), can 
win billions of brains in a single swoop, a process called 
{\it brain condensation} \cite{ZSR}.
The associated profits go with some square root of this 
condensation ratio, as reflected by the four-five orders of magnitude
gap between the salary of top executives versus their 
least paid employees \cite{WEA}. 
It is perhaps no coincidence to hear Zuckerberg proclaiming 
that "connectivity is a human right". 
The next interesting step is to realize that steering the "sentiments" of
human beings amounts to controlling a comparatively small number
of high-level "psychological variables", a piece of 
cake for the most powerful
ML applications, with little or no need of Insight \cite{ZSR}.

Trying to cure Alzheimer with ML shows a very different movie, one where  
CoD hits hard, Correlation cannot replace Causation, 
Control cannot replace Insight, a ballroom called Science \cite{VU}.
Yet, the tools are the same, it is basically the 
very same machinery described by the equation (\ref{ML}) above!  
That's why Science is so easily served as a strawman for Control.

This is where AAI becomes as dangerous as never before in the history
of science: it just trickles into our habits, step by step, no trauma, 
no shock, a silent but relentless conquer of our brain towards the ZSR goal. 
And please, make no mistake, the main item on sale with ZSR is not the 
commercial product but our brain instead. 

The motto {\it Why learn if you can look it up?} speaks loud for the above.

So, while science eventually gets important contributions 
from ML, if only hardly in proportion to bombastic claims,
sky-rocketing sales behind the scenes is a sure thing.
CS Lewis says it best: 
"When men over-emphasize the control over nature as scientific 
progress, what they really mean is control of a few 
human beings over many other human beings".

\section{A call to algorethics}

What to do then? 

In the recent years there have been increasing calls to 
"Algorethics", the idea being to inject "ethical constraints" 
into the ML algorithms \cite{BEN}.  
This is certainly a commendable goal and, as nicely 
discussed in Kearn and Roth's book \cite{KEA}, quite likely
a technically doable one as well. 

Question is: will it really fix the ZSR issue?

I sincerely doubt it, no matter how good the law, history
shows that the smart villain always manages to find the next loophole.
More effective, I think, is to pursue the inherent 
spiritual drive of Science (and capital R Religion, of course), 
namely the pursuit of Insight for the pleasure of 
finding things out, period.
The rest will follow, it always did.

\section{Summary}
Machine learning has potential to provide
a substantial new entry to the world of 
scientific investigation, as it can occasionally 
capture levels of complexity beyond reach 
for our best theoretical and computational models.
However the blind pursuit of ever-increasing number
of weights with little or no attention to Insight 
versus Control, is overly dangerous in several respects. 
Besides being energetically unsustainable, it paves the way to
zero-sales-resistence strategies whereby Science is served as
a perfect strawman for Control and financial power.  

\section{Acknowledgements}
The author is grateful to SISSA for financial support under the
"Collaborations of Excellence" initiative as well as 
to the Simons Foundation for supporting several enriching visits of his.
He also wishes to acknowledge many enlightening discussions 
over the years with PV Coveney, A. Laio and D. Spergel.
I am also grateful to M. Durve for critical reading of the manuscript.

\end{document}